\documentclass[letterpaper, 10 pt, conference]{ieeeconf}  
\IEEEoverridecommandlockouts
\usepackage{comment}
\usepackage{cite}
\newcommand{\eat}[1]{}
\usepackage{booktabs}
\usepackage{multicol}
\usepackage{multirow}
\usepackage{diagbox}
\usepackage{amsmath}
\newcolumntype{K}[1]{>{\centering\arraybackslash}p{#1}}

\usepackage{color}
\usepackage [autostyle, english = american]{csquotes}
\MakeOuterQuote{"}
\usepackage{xcolor}
\let\labelindent\relax 
\usepackage{enumitem}

\usepackage[symbol]{footmisc}

\usepackage{mathtools}

\usepackage{amsmath}
\usepackage{amstext}
\usepackage{amssymb}
\usepackage{amsfonts}
\usepackage{float}

\usepackage{amsthm}  
\usepackage[normalem]{ulem} 

\usepackage{subcaption}
\usepackage{caption}
\usepackage{todonotes}

\makeatletter

\newcommand{\Rmnum}[1]{\expandafter\@slowromancap\romannumeral #1@}
\usepackage{savesym}

\usepackage{algorithm}
\usepackage{algorithmicx} 
\usepackage{algpseudocode} %
\savesymbol{AND}
\usepackage[group-separator={,},group-minimum-digits={3}]{siunitx}

\usepackage{graphicx} 
\usepackage{epsfig} 

\usepackage{times} 
\usepackage{amsmath} 
\usepackage{amssymb}  
\usepackage{comment}

\makeatletter
\let\NAT@parse\undefined
\makeatother
\usepackage{hyperref}
\hypersetup{
   colorlinks=true,
    linkcolor= blue,
    allcolors=blue,
    citecolor = blue,
    filecolor=black,
    urlcolor=blue,
    }
\usepackage{mathrsfs}

\usepackage[symbol]{footmisc}

\usepackage[protrusion=true,expansion=true]{microtype}

\usepackage[extramath,probability,reviewmark]{mylatexdefs}

\title{\LARGE \bf

{Modified-Emergency Index (MEI): A Criticality Metric for Autonomous Driving in Lateral Conflict}}

\author{Hao Cheng$^{1}$, Yanbo Jiang$^{1}$, Qingyuan Shi$^{1}$, Qingwen Meng$^{1}$, Keyu Chen$^{1}$, Wenhao Yu$^{1}$,\\ Jianqiang Wang$^{1}$, Sifa Zheng*$^{1}$
\thanks{This work was supported by Tsinghua-Toyota Joint Research Institute Inter-disciplinary Program.}
\thanks{$^{1}$Hao Cheng, Yanbo Jiang, Qingyuan Shi,  Qingwen Meng, Keyu Chen, Wenhao Yu, Jianqiang Wang, Sifa Zheng are with School of Vehicle and Mobility, Tsinghua University, Beijing 100084, P.~R.~China.}
\thanks{Corresponding author: Sifa Zheng, E-mail: zsf@tsinghua.edu.cn}
}

\begin{document}

\maketitle
\thispagestyle{empty}
\pagestyle{empty}

\begin{abstract}

Effective, reliable, and efficient evaluation of autonomous driving safety is essential to demonstrate its trustworthiness. Criticality metrics provide an objective means of assessing safety. However, as existing metrics primarily target longitudinal conflicts, accurately quantifying the risks of lateral conflicts—prevalent in urban settings—remains challenging. This paper proposes the Modified-Emergency Index (MEI), a metric designed to quantify evasive effort in lateral conflicts. Compared to the original Emergency Index (EI), MEI refines the estimation of the time available for evasive maneuvers, enabling more precise risk quantification. We validate MEI on a public lateral conflict dataset based on Argoverse-2, from which we extract over 1,500 high-quality AV conflict cases, including more than 500 critical events. MEI is then compared with the well-established ACT and the widely used PET metrics. Results show that MEI consistently outperforms them in accurately quantifying criticality and capturing risk evolution. Overall, these findings highlight MEI as a promising metric for evaluating urban conflicts and enhancing the safety assessment framework for autonomous driving. The open-source implementation is available at \url{https://github.com/AutoChengh/MEI}.

\end{abstract}

\section{Introduction}


Autonomous driving technology holds great promise for improving traffic safety, efficiency, and sustainability. To ensure the reliability of autonomous vehicles (AVs), it is crucial to establish trustworthy and effective safety metrics. A proactive, objective approach is to develop and apply criticality metrics, which quantify “the composite risk involving all actors as the traffic situation evolves” \cite{westhofen2023criticality}. These metrics serve as surrogate measures of risk from specific perspectives and are widely used for AV safety evaluation, identifying critical events to build scenario libraries, and supporting regulatory frameworks.

Urban environments present numerous high-risk conflicts, making them a primary focus in developing autonomous driving systems. These settings often involve multi-angle conflicts like intersection crossing and merging, frequent traffic violations, and complex interactions with various Vulnerable Road Users (VRUs). Thus, accurately assessing AV safety in urban contexts is a core task for criticality metrics.

Most criticality metrics trace back to Surrogate Safety Measures (SSMs) from the 1960s–70s, designed as proactive, efficient alternatives to crash data \cite{tafidis2023application}. Classic SSMs such as Time-to-Collision (TTC) remain valuable for longitudinal risks like rear-end collisions, but are limited for the lateral conflicts common in cities. Metrics such as Post-Encroachment Time (PET) \cite{allen1978analysis} and Time Advantage (TAdv) \cite{laureshyn2010evaluation} were proposed to address this gap.

Further, many efforts extend TTC to more complex 2D interactions, exemplified by Extended TTC (ETTC) \cite{ward2015extending}, Anticipated Collision Time (ACT) \cite{venthuruthiyil2022anticipated}, and 2D-TTC \cite{guo2023modeling}. Recently, the Emergency Index (EI) \cite{cheng2025emergency} was proposed to quantify evasive difficulty in multi-angle conflicts. EI builds on Interaction Depth (InDepth), measuring how much two vehicles intrude spatially if no avoidance occurs—a positive InDepth implies an unavoidable collision without evasive action. The task is to reduce InDepth from positive to negative within the remaining time, with EI capturing this by defining the required rate of change. While InDepth fully accounts for vehicle geometry, EI estimates remaining time via a centroid-based approach, which may limit accuracy in critical near-miss scenarios. Moreover, despite several metrics for lateral conflicts, few studies have rigorously validated their effectiveness~\cite{ismail2011methodologies, salamati2011development, arun2021systematic, jiao2025unified}. Understanding each metric’s strengths and failures is essential for reliable lateral conflict assessment.

In this paper, we propose a refined criticality metric, the Modified-Emergency Index (MEI), to address the challenge of accurately evaluating safety-critical scenarios in urban environments. We validate MEI on an open-source lateral conflict dataset derived from Argoverse-2 and compare it against two widely used metrics, ACT and PET.

The contributions of this paper include:

\begin{enumerate}
\item We propose MEI, a novel criticality metric tailored for lateral conflict scenarios, which extends the previously introduced EI by more accurately estimating temporal proximity. This enhancement enables a more precise evaluation of evasive effort.

\item By analyzing over 1,500 AV-involved lateral interaction events from the Argoverse-2 dataset, we find that MEI exhibits clear advantages over existing metrics (e.g., ACT and PET) in both risk assessment accuracy and capturing risk evolution trends. This suggests that MEI may be a better choice for evaluating the criticality of complex lateral conflicts in urban scenarios.


\end{enumerate}

The rest of the paper is organized as follows. Section~\ref{sec:methodology} details the computational methodology of MEI. Section~\ref{sec:results} presents validation results and failure analyses of MEI and other metrics. Finally, Section~\ref{sec:conclu} concludes the paper.

\section{Methodology}
\label{sec:methodology}

In this section, we present the computation of the Modified-Emergency Index (MEI), detailing the calculations of Interaction Depth (InDepth) and Time for Evasive Maneuver (TEM). We also formalize criteria within MEI to identify potential conflicts, critical events, and crashes. Finally, we describe the Lateral Conflict Resolution Dataset and our procedure for extracting high-risk cases.

\subsection{Interaction Depth (InDepth)}
We assume that the state vector of vehicle \( i \) at a given time \( t \in \mathbb{R}^+ \) is denoted as  
\[
 \boldsymbol{S}_i(t) = [x_i(t),\ y_i(t),\ v_i(t),\ \theta_i(t),\ l_i,\ w_i]^T,
\]
where \( x_i(t) \) and \( y_i(t) \) represent the X and Y coordinates of the geometric center of vehicle \( i \),  
\( v_i(t) \) and \( \theta_i(t) \) denote its speed magnitude and heading angle,  
and \( l_i \) and \( w_i \) represent the length and width of the vehicle, respectively.  
Based on this definition, the position vector \( \boldsymbol{P}_i(t) \), velocity vector \( \boldsymbol{v}_i(t) \), and direction unit vector \( \boldsymbol{\theta}_i(t) \) can be expressed as follows:

\begin{equation}
\left\{
\begin{aligned}
    \boldsymbol{P}_i(t) &= \begin{bmatrix} x_i(t),\ y_i(t) \end{bmatrix}^T \\
    \boldsymbol{v}_i(t) &= \begin{bmatrix} v_i(t)\cos\left(\theta_i(t)\right),\ v_i(t)\sin\left(\theta_i(t)\right) \end{bmatrix}^T \\
    \boldsymbol{\theta}_i(t) &= \begin{bmatrix} \cos\left(\theta_i(t)\right),\ \sin\left(\theta_i(t)\right) \end{bmatrix}^T 
    = \frac{\boldsymbol{v}_i(t)}{\|\boldsymbol{v}_i(t)\|}
\end{aligned}
\right.
\end{equation}


In the relative coordinate system of vehicle $B$, the relative position vector of vehicle $A$ is denoted as $\boldsymbol{P}_{AB} = \begin{bmatrix} x_A - x_B,\ y_A - y_B \end{bmatrix}^T$, and the relative velocity vector is computed as:

\begin{equation}
    \boldsymbol{v}_{AB} =\boldsymbol{v}_A - \boldsymbol{v}_B = 
    \begin{bmatrix}
        v_A \cos(\theta_A) - v_B \cos(\theta_B) \\
        v_A \sin(\theta_A) - v_B \sin(\theta_B)
    \end{bmatrix}
    \label{eq:relative_velocity}
\end{equation}

The direction unit vector of the relative velocity is:

\begin{equation}
    \boldsymbol{\theta}_{AB} = \frac{\boldsymbol{v}_{AB}}{\|\boldsymbol{v}_{AB}\|}
    \label{eq:theta_ab}
\end{equation}

The tangential Euclidean distance between the two centers of mass is given by:

\begin{equation}
    D_c^T = \left\| \boldsymbol{P}_{AB} \times \boldsymbol{\theta}_{AB} \right\|
    \label{eq:distance_tangent}
\end{equation}

To account for vehicle size, we define the vectors from the vehicle center to the four corners of the rectangular body as:

\begin{equation}
\left\{
\begin{aligned}
    \overrightarrow{AA_{i}} &= (-1)^{\left\lfloor i - \frac{1}{2} \right\rfloor} \frac{l_A}{2} \boldsymbol{\theta}_A + (-1)^i \frac{w_A}{2} \boldsymbol{\theta}_A^{\perp}, \quad i \in \{1,2,3,4\} \\
    \overrightarrow{BB_{j}} &= (-1)^{\left\lfloor j - \frac{1}{2} \right\rfloor} \frac{l_B}{2} \boldsymbol{\theta}_B + (-1)^j \frac{w_B}{2} \boldsymbol{\theta}_B^{\perp}, \quad j \in \{1,2,3,4\}
\end{aligned}
\right.
\end{equation}

Here, \( \boldsymbol{\theta}_A^\perp \) is obtained by rotating \( \boldsymbol{\theta}_A \) counterclockwise by 90°, following the right-hand rule.

To account for vehicle size, we define the projection radii 
$d_A$ and $d_B$ as the maximum distance from the vehicle center to its body corners orthogonal to the relative velocity:



\begin{equation}
\left\{
\begin{aligned}
    d_A &= \max_{i \in \{1,2,3,4\}} \left\| \overrightarrow{AA_{i}} \times \boldsymbol{\theta}_{AB} \right\| \\
    d_B &= \max_{j \in \{1,2,3,4\}} \left\| \overrightarrow{BB_{j}} \times \boldsymbol{\theta}_{AB} \right\|
\end{aligned}
\right.
\end{equation}

\begin{figure}[htbp]
    \centering
    \includegraphics[width=0.9\linewidth]{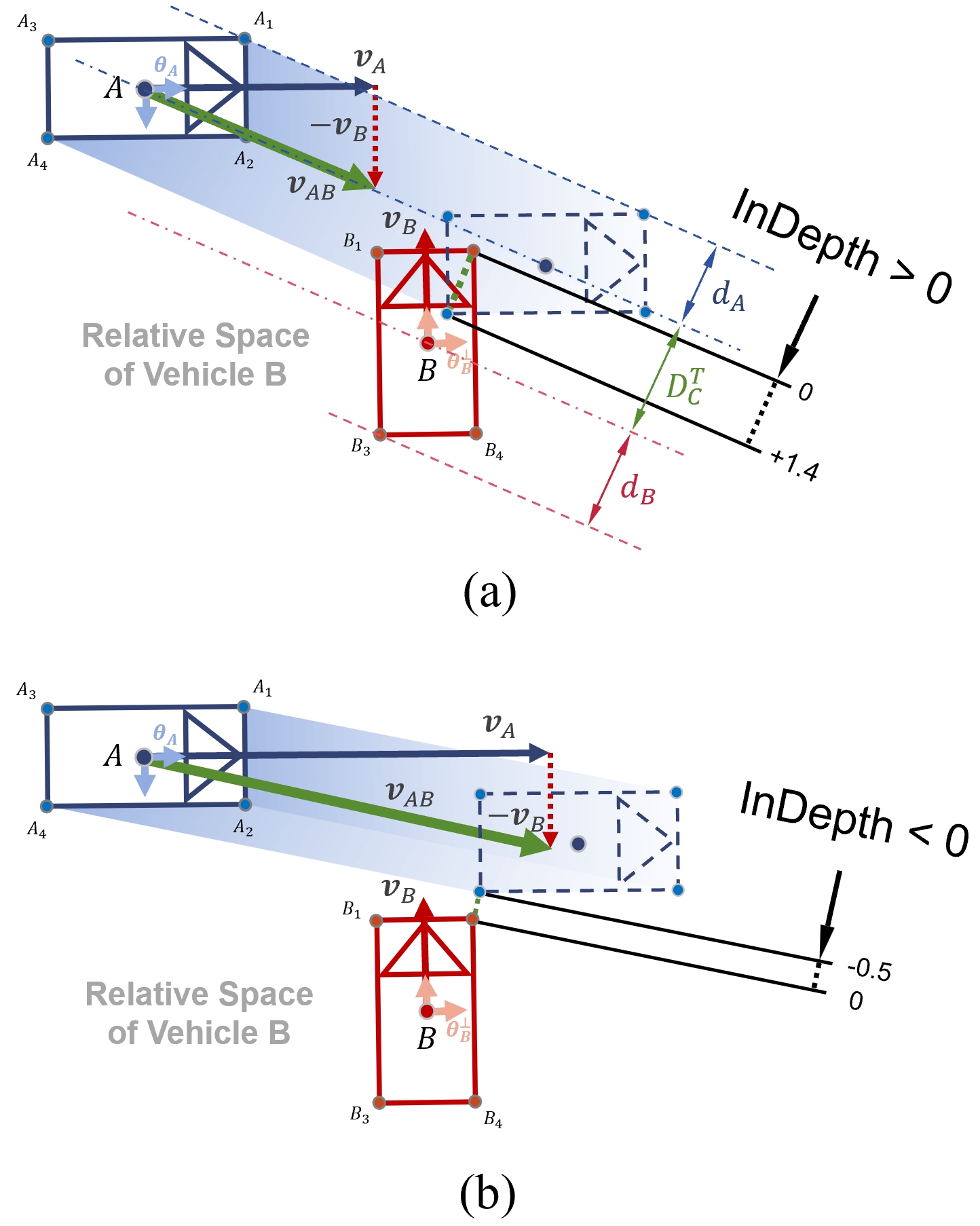}
    \caption{Illustration of InDepth. (a) When InDepth is positive, the vehicles will collide unless either (or both) vehicle takes effective evasive action. (b) When InDepth is negative, they won’t collide even without any evasive action.}
    \label{fig:InDepth}
\end{figure}

\begin{figure}[htbp]
    \centering
    \includegraphics[width=0.95\linewidth]{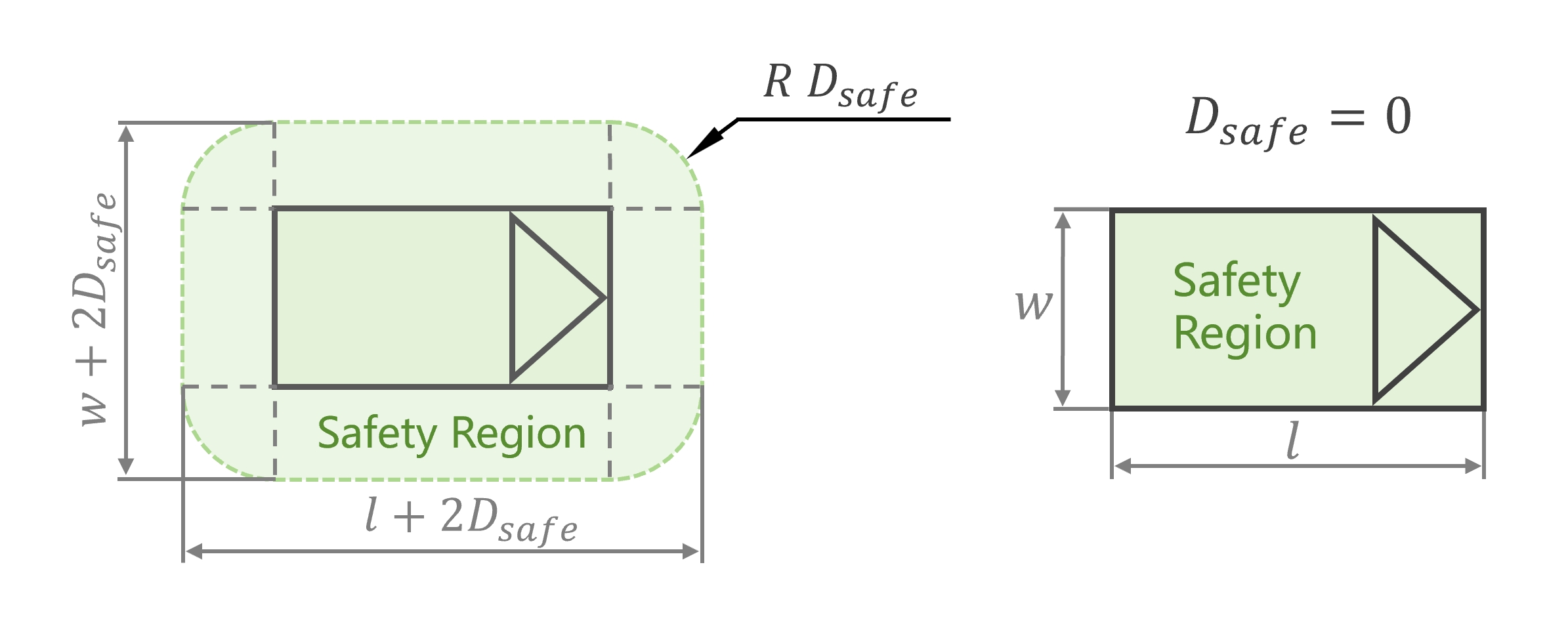}
    \caption{Illustration of the safety region. Specifically, when $D_{\text{safe}} = 0$, the safety region is identical to the vehicle body.}
    \label{fig:Dsafe}
\end{figure}

Due to vehicles’ risk-sensitive nature, the space they aim to protect often extends beyond the physical body of the vehicle and may include a surrounding buffer zone~\cite{hou2014new}. We model this protected area using a safety region, as illustrated in Fig.~\ref{fig:Dsafe} The safety region is defined as a rounded rectangle enclosing the vehicle body, with a corner radius denoted as $D_{\text{safe}}$. Here, $D_{\text{safe}}$ can be interpreted as the minimum inter-vehicle distance acceptable to the vehicle.

Furthermore, we introduce the concept of Interaction Depth (InDepth) to quantify the maximum depth to which two vehicles may intrude into each other’s safety regions in the future. The calculation is given by:

\begin{equation}
\textit{InDepth} = d_A + d_B - D_c^T + D_{\text{safe}}
\label{eq:indepth}
\end{equation}


In particular, when $D_{\text{safe}} = 0$, the safety region coincides with the vehicle body size, implying no safety redundancy. For consistency and generality, we set $D_{\text{safe}} = 0$ in the subsequent analyses.

\subsection{Modified-Emergency Index (MEI)}

For two vehicles in a conflict state, a transition to a safer state requires the execution of an evasive maneuver within a limited time window, referred to as the Time for Evasive Maneuver (TEM). The necessary change for avoiding collision is measured by the Interaction Depth (InDepth). We define the Modified-Emergency Index (MEI) as the ratio between InDepth and TEM, as shown in Eq.~\ref{eq:mei}. A smaller TEM or a larger InDepth results in a higher MEI, indicating that MEI is positively correlated with both the risk level and the urgency of the conflict.

\begin{equation}
    \textit{MEI} = \frac{\textit{InDepth}}{\textit{TEM}}
    \label{eq:mei}
\end{equation}

\begin{figure}[htbp]
    \centering
    \includegraphics[width=0.95\linewidth]{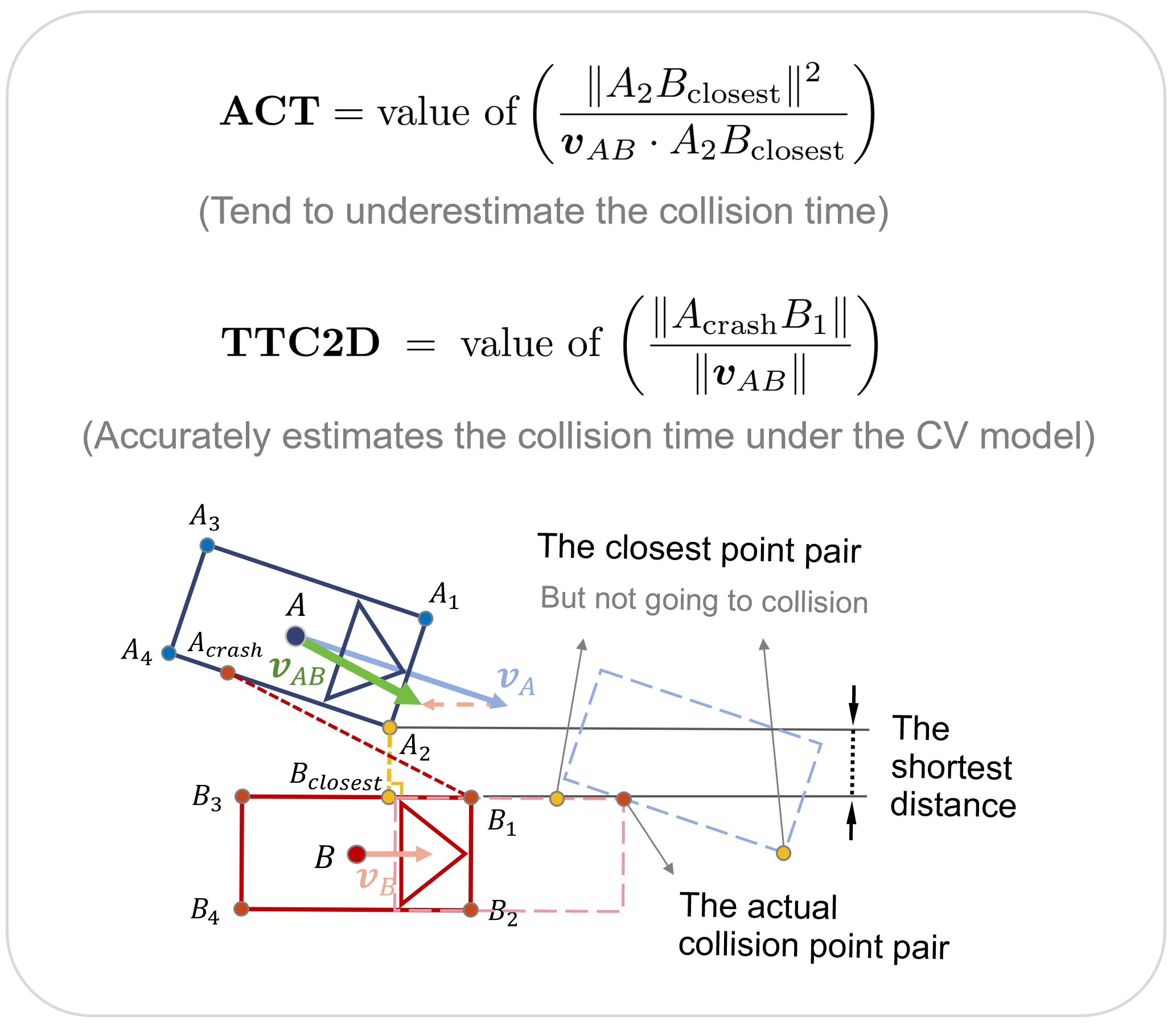}
    \caption{Illustration of ACT and TTC2D calculations. Notably, the closest point pair identified by ACT may not be the actual pair expected to collide, leading it to estimate a collision time nearly half that of TTC2D in this case.}
    \label{fig:TEM}
\end{figure}

To quantify TEM, we adopt the TTC2D proposed by Jiao~\cite{jiao2023ttc}. Unlike ACT, which considers the closest pair of points on the vehicles’ bounding boxes and estimates the time to collision for these nearest points, TTC2D evaluates the actual point pair expected to collide under the Constant Velocity (CV) model. As shown in Fig.~\ref{fig:TEM}, TTC2D can be regarded as an exact evaluation of collision time under the CV assumption.\footnotemark

\footnotetext[1]{This does not mean ACT’s logic is inferior; its conservatism may be beneficial in  certain contexts, but this lies beyond our scope.}

Furthermore, by incorporating the Conflict Detection Model (CDM) proposed by Cheng et al.~\cite{cheng2025emergency} and setting a reasonable threshold for TEM (e.g., $\text{TEM}^*=3\,\text{s}$), we can formally classify conflicts at any given time into four risk levels: Non-Conflict, Potential Conflict, Critical Conflict, and Crash, as summarized in Table~\ref{tab:Judging Conflict}.

\begin{table*}[t]
\centering
\caption{The MEI framework for judging potential conflict, critical conflict, and crash moment}
\label{tab:Judging Conflict}
\renewcommand{\arraystretch}{1.1}
\begin{tabular}{p{2.8cm} >{\centering\arraybackslash}p{2.8cm} >{\centering\arraybackslash}p{2.8cm} >{\centering\arraybackslash}p{2.8cm} >{\centering\arraybackslash}p{2.8cm}} 
\toprule
 & Non-Conflict & Potential Conflict & Critical Conflict & Crash \\
\midrule
Condition $Q$ in CDM \cite{cheng2025emergency} & $\times$ & $\checkmark$ & $\checkmark$ & $\checkmark$ \\
\cmidrule{1-5}
TEM (s) & - & - & $\leq \text{TEM}^*$ & $\rightarrow 0$ \\
\cmidrule{1-5}
InDepth (m) & - & - & $\geq 0$ & $\geq D_{\text{safe}}$ \\
\cmidrule{1-5}
MEI (m/s) & - & - & $\geq 0$ & $\rightarrow +\infty$ \\
\bottomrule
\end{tabular}
\end{table*}

\subsection{Lateral Conflict Resolution Dataset}

The Argoverse-2 dataset~\cite{wilson2023argoverse}, collected by an AV fleet in six cities, contains 250,000 driving scenarios with a focus on safety-critical and long-tail events. Each scenario spans 11 seconds sampled at 10 Hz, along with high-definition maps including vectorized lanes and drivable area annotations. Based on this, Li et al.~\cite{li2024lateral} construct and release a high-quality lateral conflict resolution dataset\footnotemark, developed through rigorous processing with anomaly correction and balanced traffic conditions across conflict types.

Using this lateral conflict dataset, we apply the Separating Axis Theorem (SAT)~\cite{gottschalk1996separating} to filter out “collision” cases caused by noise and errors. This yields 1,548 conflict instances with maximum MEI $>0$ involving the AV. As summarized in Table~\ref{tab:Judging Conflict}, 501 cases (32.4\%) are classified as critical conflicts and 1,047 (67.6\%) as potential conflicts.

\footnotetext[2]{\url{https://github.com/RomainLITUD/conflict_resolution_dataset}}

\section{Results}
\label{sec:results}


In this section, we analyze high-risk AV cases using the lateral conflict dataset introduced in Section~\ref{sec:methodology}. We apply MEI along with widely used metrics such as ACT and PET. We first present statistical results, then provide case studies that demonstrate the advantages of MEI.
\subsection{Statistical Analysis of AV Conflicts Using Criticality Metrics}



\begin{figure}[ht]
    \centering
    \includegraphics[width=0.95\linewidth]{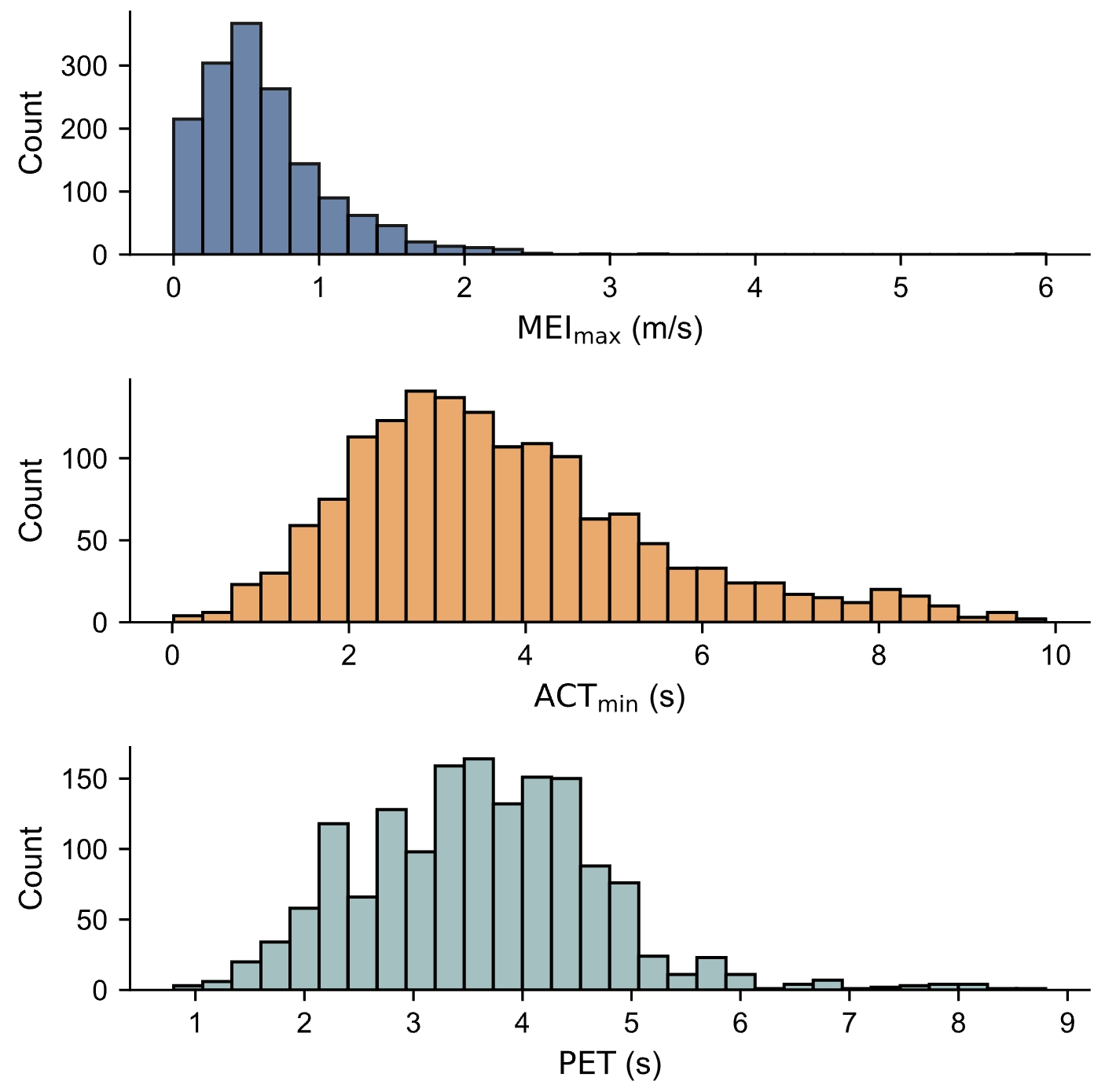}
    \caption{Distributions of criticality metrics: (top) MEI\textsubscript{max}, (middle) ACT\textsubscript{min}, and (bottom) PET.}
    \label{fig:count}
\end{figure}

To further investigate the risk distribution of AV-involved conflict events, we compute the maximum MEI, the minimum ACT, and the PET for each of the 1,548 conflict samples. The distributions of these criticality metrics are illustrated in Fig.~\ref{fig:count}. Furthermore, Table~\ref{tab:risk_levels} summarizes the risk-level thresholds based on percentile rankings. These statistical results provide a foundation for stratifying conflict severity and support the establishment of evaluation benchmarks for identifying critical events. In the following analysis, we identify potential failure cases by examining instances where the risk levels differ significantly across different criticality metrics.

\begin{table}[ht]
\renewcommand{\arraystretch}{1.3} 
\centering
\caption{Risk-level thresholds for {MEI\textsubscript{max}}, {ACT\textsubscript{min}}, and {PET} from the Argoverse-2 lateral conflict dataset}
\label{tab:risk_levels}
\begin{tabular}{lccc}
\toprule
\textbf{Risk Level} & \textbf{MEI\textsubscript{max} (m/s)} & \textbf{ACT\textsubscript{min} (s)} & \textbf{PET (s)} \\
\midrule
\textbf{Top 1\%}   & \textbf{2.13} (99th)  & \textbf{0.77} (1st)  & \textbf{1.40} (1st) \\
\textbf{Top 5\%}   & \textbf{1.52} (95th)  & \textbf{1.43} (5th)  & \textbf{2.00} (5th) \\
\textbf{Top 10\%} & \textbf{1.22} (90th)  & \textbf{1.85} (10th) & \textbf{2.20} (10th) \\
\textbf{Top 25\%}  & \textbf{0.81} (75th)  & \textbf{2.52} (25th) & \textbf{2.80} (25th) \\
\midrule
Top 50\%  & 0.53 (50th)  & 3.47 (50th) & 3.60 (50th) \\
Top 75\%  & 0.33 (25th)  & 4.66 (75th) & 4.30 (75th) \\
Top 90\%  & 0.14 (10th)  & 6.19 (90th) & 4.80 (90th) \\
Top 95\%  & 0.08 (5th)   & 7.43 (95th) & 5.30 (95th) \\
Top 99\%  & 0.01 (1st)   & 8.78 (99th) & 6.95 (99th) \\
\bottomrule
\end{tabular}
\end{table}


\subsection{Failure Case Analysis}

\subsubsection{\textbf{Case 1: High-Risk Near-Miss Scenario}}
\begin{figure*}[t]
    \centering
    \includegraphics[width=0.95\linewidth]{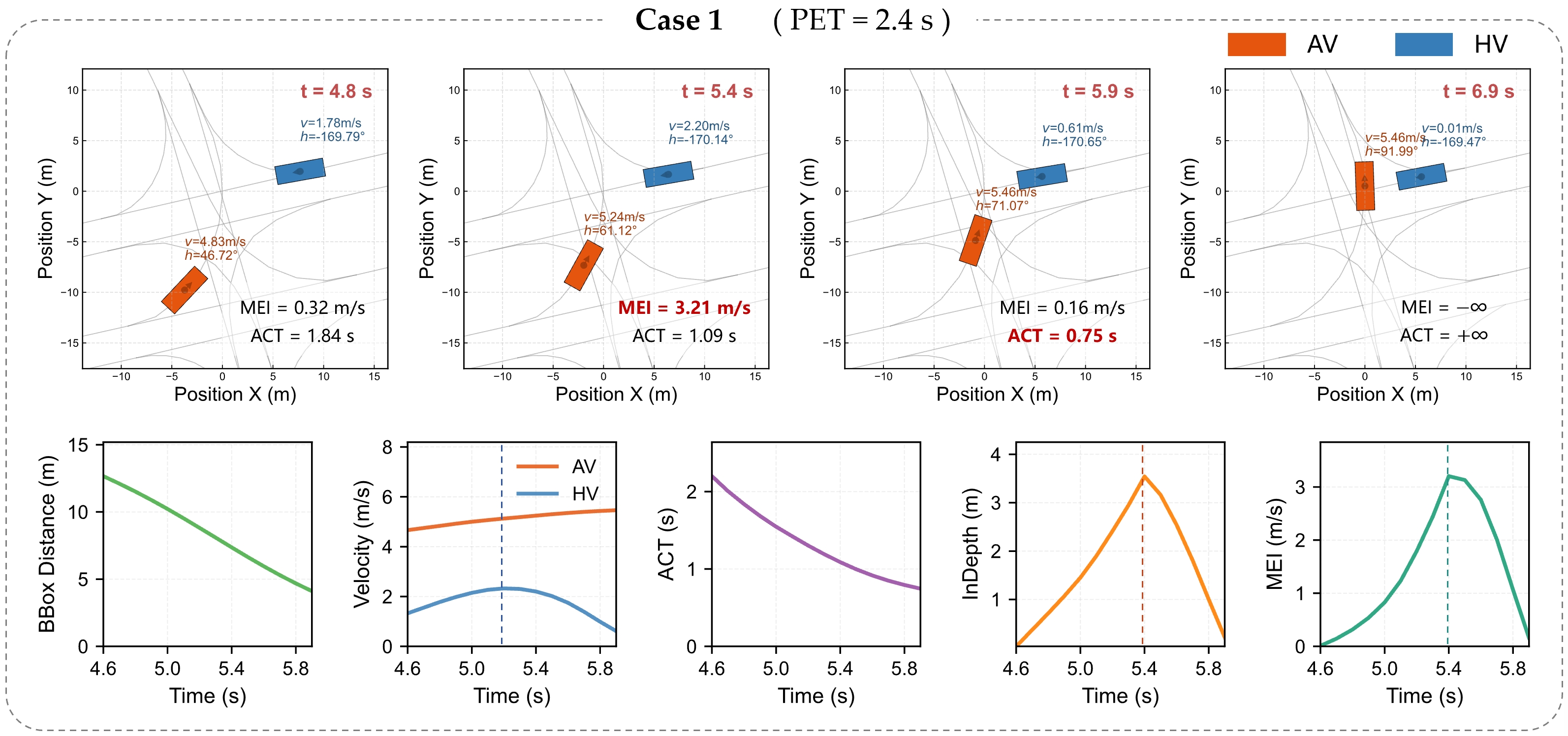}
    \caption{A lateral conflict where the AV turns left and encounters a straight-driving HV. MEI’s risk assessment aligns with vehicle behavior, suggesting higher frame-level accuracy.}
    \label{fig:case4}
\end{figure*}

As shown in Fig.~\ref{fig:case4}, Case~1 involves a lateral conflict in which the AV makes a left turn at an intersection and encounters a straight-moving HV. In this case, PET is $2.4\,\text{s}$. The maximum MEI is $3.21\,\text{m/s}$ (at $t=5.4\,\text{s}$), and the minimum ACT is $0.75\,\text{s}$ (at $t=5.9\,\text{s}$), both indicating a high-risk safety-critical event. The risk arises from the HV accelerating between $t=4.6\,\text{s}$ and $5.2\,\text{s}$ to compete for right-of-way, then decelerating after $t=5.2\,\text{s}$ to yield. MEI captures this rise-and-fall pattern, reflecting the HV’s acceleration followed by deceleration. By contrast, ACT steadily decreases from $2.20\,\text{s}$ to $0.75\,\text{s}$ over this period, indicating a continuously increasing risk that does not align with the actual criticality evolution. This demonstrates that MEI more precisely tracks changing risk than ACT, with better frame-level responsiveness, highlighting its potential for real-time risk monitoring and vehicle warning systems.


\subsubsection{\textbf{Case 2: False Alarm (False Positive) by ACT}}
\begin{figure*}[t]
    \centering
    \includegraphics[width=0.95\linewidth]{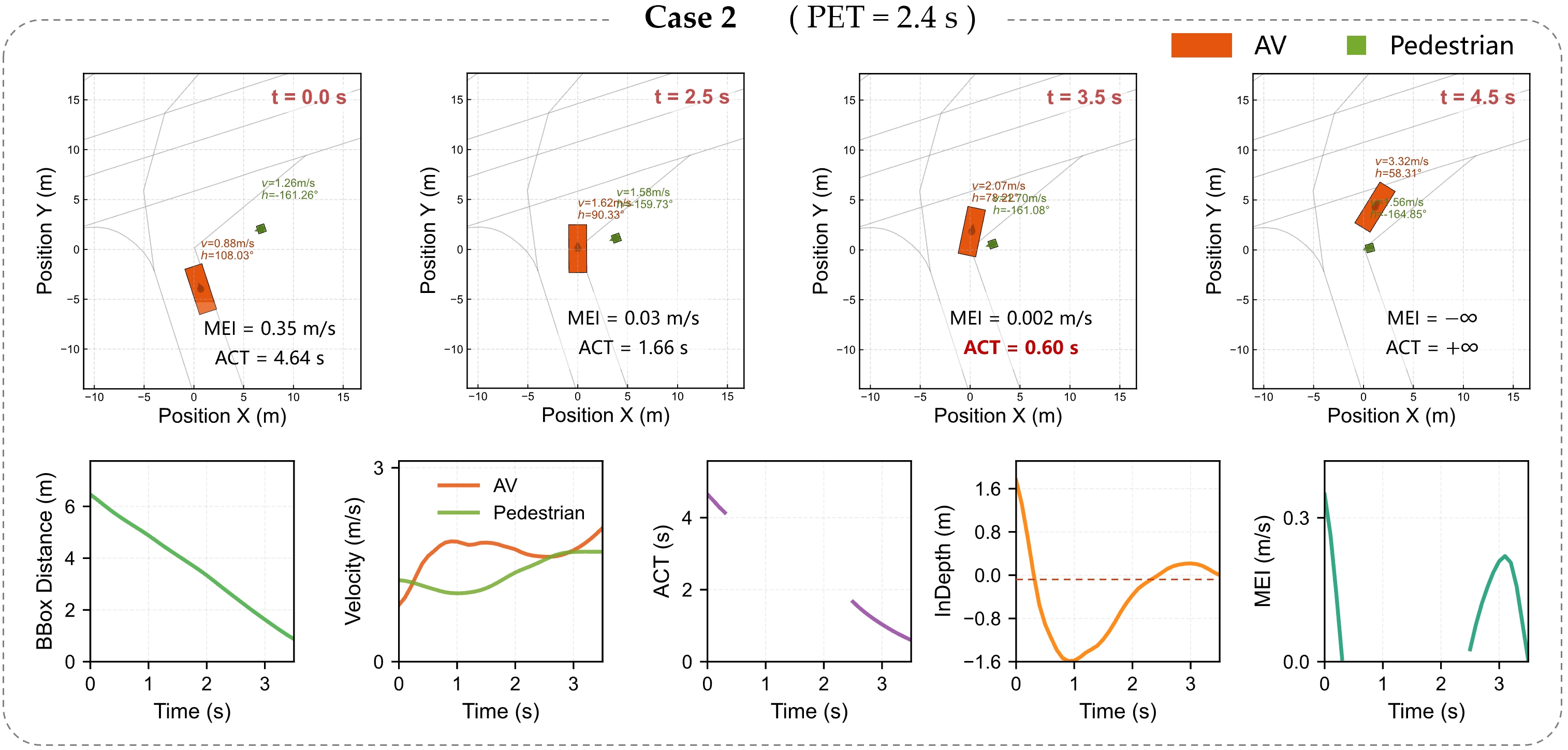}
    \caption{A lateral conflict scenario where the AV turns right and a pedestrian approaches from the right. The minimum ACT is $0.60\,\text{s}$, indicating that ACT may overestimate the actual risk.}
    \label{fig:case1}
\end{figure*}

As shown in Fig.~\ref{fig:case1}, this case involves a lateral conflict where the AV turns right at an intersection and encounters a pedestrian approaching from the right. Throughout the scenario, the PET is $2.4\,\text{s}$. In this case, the maximum MEI is $0.35\,\text{m/s}$ (occurring at $t=0.1\,\text{s}$), indicating that the situation is a manageable safety-critical event. However, at $t = 3.5\,\mathrm{s}$, ACT reaches a minimum of $0.60\,\mathrm{s}$ (indicating top $1\%$ risk), even though the AV is about to leave the conflict zone and the pedestrian is preparing to pass behind it. For MEI, although TEM at this moment is only $0.60\,\mathrm{s}$, InDepth is merely $0.0014\,\mathrm{m}$, resulting in a negligible MEI of $0.0024\,\mathrm{m/s}$, which indicates low actual evasive difficulty.

\subsubsection{\textbf{Case 3: Missed Detection by PET}}

\begin{figure*}[t]
    \centering
    \includegraphics[width=0.95\linewidth]{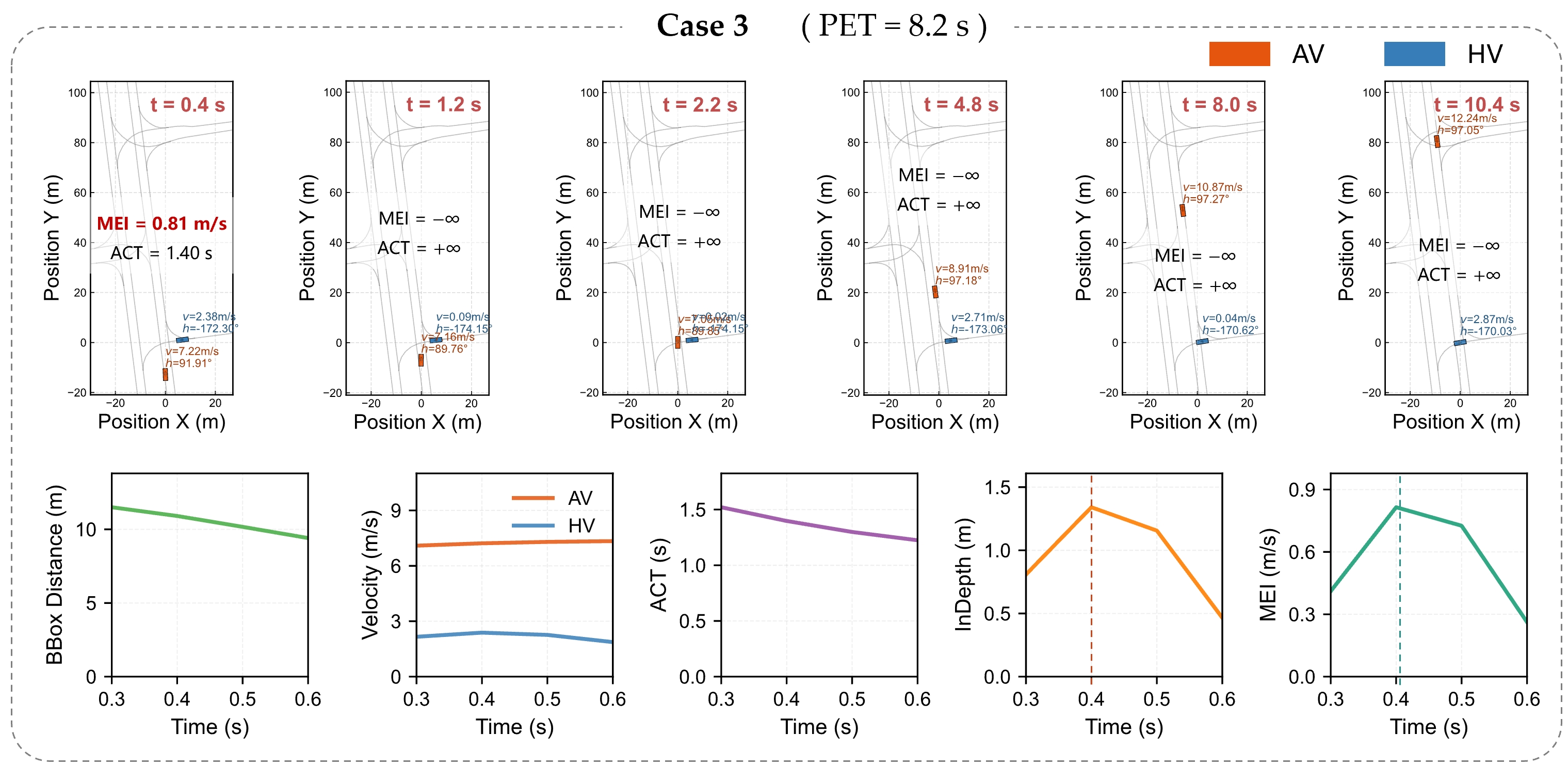}
    \caption{A lateral conflict where the AV proceeds straight and encounters an HV from the right. Despite a PET of $8.2\,\text{s}$, both ACT and MEI indicate a safety-critical event, suggesting that PET may underestimate the risk.}
    \label{fig:case2}
\end{figure*}

As shown in Fig.~\ref{fig:case2}, Case~3 involves a lateral conflict where the AV proceeds straight through an intersection and encounters a Human-driven Vehicle (HV) approaching from the right. The maximum MEI is $0.81\,\text{m/s}$ (occurring at $t=0.4\,\text{s}$), and the minimum ACT is $1.22\,\text{s}$ (at $t=0.6\,\text{s}$), both indicating a safety-critical event. However, at $t=2.2\,\text{s}$, after the AV has passed through the intersection, the HV does not proceed immediately but instead enters the potential conflict zone at $t=10.4\,\text{s}$. This results in a PET value of $8.2\,\text{s}$, which suggests a non-conflict or only a potential conflict. The delayed entry of the HV may be due to a traffic signal or an unexpected situation (the dataset does not specify the cause). As a result, PET appears to underestimate the actual risk in this case.

\subsubsection{\textbf{Case 4: False Alarm (False Positive) by PET}}

\begin{figure*}[t]
    \centering
    \includegraphics[width=0.95\linewidth]{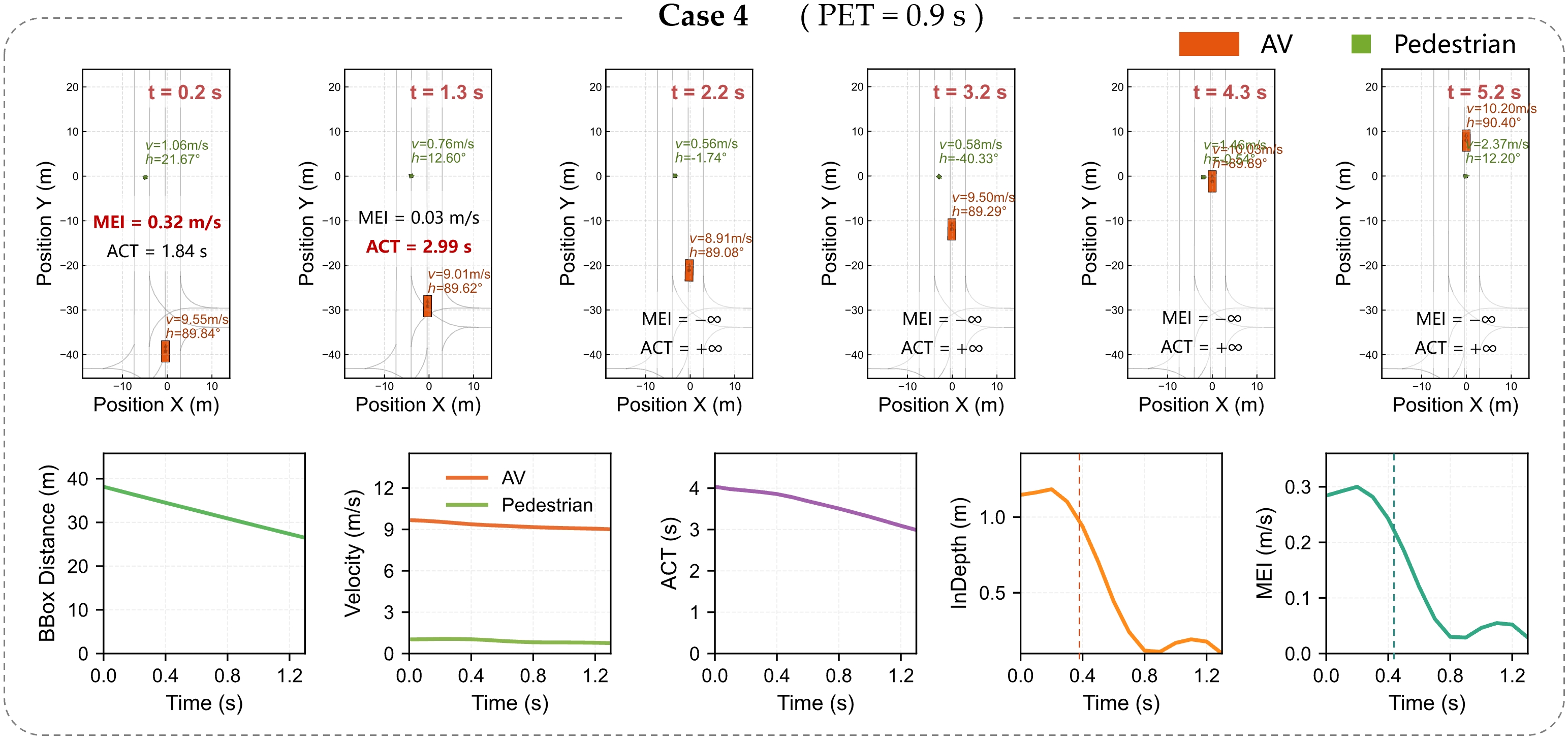}
    \caption{A lateral conflict where the AV proceeds straight and encounters a pedestrian crossing from the left. The PET is $0.9\,\text{s}$, but ACT and MEI indicate low and manageable risk, suggesting that PET may overestimate the actual risk.}
    \label{fig:case3}
\end{figure*}

As shown in Fig.~\ref{fig:case3}, Case~4 involves a lateral conflict where the AV proceeds straight and encounters a pedestrian crossing the road from the left. The maximum MEI is $0.32\,\text{m/s}$ (at $t=0.2\,\text{s}$) and the minimum ACT is $2.99\,\text{s}$ (at $t=1.3\,\text{s}$), both indicating low, manageable risk. However, after the AV exits the conflict zone at $t=4.3\,\text{s}$, the pedestrian enters at $t=5.2\,\text{s}$, resulting in a PET of only $0.9\,\text{s}$, placing it in the top 1\% risk level under PET’s threshold. Further analysis shows the pedestrian enters only after the AV has passed and maintains constant speed, suggesting no actual danger. PET thus overestimates the risk in this case.


Cases~2 and~3 collectively indicate that PET’s risk quantification is strongly influenced by the behavior of the second road user. When the critical interaction primarily occurs during the lead vehicle’s traversal, PET may not reliably capture the actual risk level.

\section{Conclusions}
\label{sec:conclu}
This paper proposes a criticality metric, the Modified-Emergency Index (MEI), to address the challenge of accurately assessing criticality in urban lateral conflict scenarios. We validate MEI using a lateral conflict resolution dataset based on Argoverse-2 and compare it with several mainstream criticality metrics. Experimental results show that MEI consistently quantifies risk and captures the trend of risk evolution accurately, indicating its strong potential as a reliable metric for assessing criticality in complex urban lateral conflicts.

This study also has some limitations. First, it mainly adopts a case study approach; future work should conduct quantitative experiments on accident correlation to further validate the metric’s effectiveness. Second, MEI is only tested in urban lateral conflicts here, and future studies could extend the analysis to highway scenarios to examine its cross-scenario applicability. Finally, while a comprehensive evaluation of autonomous driving should consider safety, efficiency, and comfort, such integrated assessment is beyond this paper’s scope but remains worth exploring in future research.


\bibliographystyle{IEEEtran.bst} 
\bibliography{reference/ref.bib}

\clearpage
\newpage
\end{document}